\newif\ifACM

\ifACM
  \documentclass[sigconf, authordraft]{acmart}
\else
  \documentclass[10pt]{article}
  \usepackage{authblk}
  \usepackage{fullpage}
\fi

\usepackage[utf8]{inputenc}

\usepackage{amsmath,amsfonts,amsthm}
\usepackage{color}
\usepackage{graphicx}
\usepackage{caption}
\usepackage{subcaption}
\usepackage{csvsimple}
\usepackage{cleveref}
\usepackage{booktabs} 

\ifACM
\setcopyright{rightsretained}

\acmDOI{}

\acmISBN{}

\acmConference[SIGSPATIAL'18]{ACM International Conference on Advances in Geographic Information Systems}{November 2018}{Seattle, Washington USA}
\acmYear{2018}
\copyrightyear{2018}

\acmPrice{xx}

\acmSubmissionID{XXX-YYY-ZZ}
\fi

\newcommand{\tobs}{t}
\newcommand{\R}{\mathbb{R}}

\begin{document}
\title{Adversarial Examples in Remote Sensing}

\ifACM
\else
\author[2]{Wojciech Czaja}
\author[1]{Neil Fendley}
\author[1]{Michael Pekala}
\author[1]{Christopher Ratto}
\author[1]{I-Jeng Wang}
\affil[1]{Johns Hopkins University Applied Physics Laboratory \protect\\ Laurel, MD \protect\\ \emph{firstname.lastname}@jhuapl.edu}
\affil[2]{University of Maryland \protect\\ Department of Mathematics \protect\\ College Park, MD \protect\\ wojtek@math.umd.edu}
\fi

\maketitle

\begin{abstract}
This paper considers attacks against machine learning algorithms used in remote sensing applications, a domain
that presents a suite of challenges that are not fully addressed by current research focused on natural image data such as ImageNet.  
In particular, we present a new study of adversarial examples in the context of satellite image classification problems.
Using a recently curated data set and associated classifier, we provide a preliminary analysis of adversarial examples in settings where the targeted classifier is permitted multiple observations of the same location over time.
While our experiments to date are purely digital, our problem setup explicitly incorporates a number of practical considerations that a real-world attacker would need to take into account when mounting a physical attack.
We hope this work provides a useful starting point for future studies of potential vulnerabilities in this setting.
\end{abstract}

\section{Introduction}
Many modern deep learning systems exhibit a lack of stability to specially-designed ``small'' perturbations to the signal input space.
While what precisely constitutes a ``small'' perturbation varies by application, it is generally understood to be a modification that leaves the (human-perceived) signal content unchanged while inducing a fundamental 
change in the output of the targeted machine learning system.
Signals designed in this manner are termed \emph{adversarial examples} (AE) , and have been a topic of substantial interest in the machine learning community recently~\cite{szegedy2013intriguing,goodfellow2014explaining,papernot2016practical,carlini2017adversarial}.

Despite widespread attention, there remain many open questions related to AE.
For example, while a number of analyses into the mathematical properties of AE have been conducted (e.g. \cite{fawzi2017robustness,faghri2018adversarial,fawzi2016robustness}), a complete theoretical understanding of this phenomenon remains elusive.
On the more applied end of the spectrum, it remains an open question how readily AE can induce errors in real-world systems under practical conditions.
This question of applicability is broad and depends upon many factors such as \emph{a priori} knowledge of the targeted system, how and where perturbations can be injected in the signal processing chain, what constraints are placed upon the perturbation, and the robustness of the adversarial examples themselves to perturbations during signal acquisition (due to changes in viewpoint, lighting, preprocessing etc.).

When AE are to be realized physically (i.e. by modifying the physical environment) the question of their robustness due to variability inherent in the sensing process  
has prompted some discussion 
regarding their practicality. 
In \cite{lu2017no}, the authors concluded that AE are not a real-world concern in the autonomous vehicle domain since many AE generated from a single anticipated perspective did not preserve their adversarial properties when perceived from other viewpoints.  
Subsequently, \cite{athalye2017synthesizing} demonstrated that, by explicitly accounting for the anticipated distribution of viewpoints (and other variations in the sensing process), it is indeed possible to construct robust AE and therefore the phenomenon merits consideration.

Machine learning for remote sensing generally concerns the detection and classification of objects on the Earth's surface and retrieving geophysical or biochemical quantities \cite{zhu2017deep}. As the proliferation of satellite imagers continues to increase, automation will be required to process the 
volumes
of data that are expected and thus may be vulnerable to AE. Remote sensing is rich setting for studying AE in part because there is often strong prior knowledge available regarding the likely sensing geometries and modalities, as well as the radiometric properties of scene elements (e.g. manmade structures, vegetation, water). Unlike natural imagery, remote sensing data are geolocated (each pixel can be associated a corresponding latitude and longitude) and are quality-controlled. While having such knowledge facilitates approaches such as \cite{athalye2017synthesizing} for generating robust AE, there are also unique challenges in this domain.
We detail a number of these challenges in this paper.


A number of studies of physically-constrained AE have been performed recently, including~\cite{kurakin2016adversarial,evtimov2017robust,athalye2017synthesizing}. 
These experiments have involved earth-based sensing in the visible spectrum at relatively close ranges, e.g. within the sensing range of the camera on an autonomous vehicle or a facial recognition system. 
Attacks against typical remote sensing modalities (e.g. satellite imagery, multi-spectral data, LIDAR, and SAR) have not yet been well-studied.
As far as the authors are aware, this work presents the first study of AE for satellite imagery in the visible spectrum. 

In this work we restrict our attention to (whole image) classification problems.
By now, deep learning has been applied to a variety of other signal processing problems that arise in remote sensing (object detection, segmentation, etc.; see \cite{zhu2017deep} for a comprehensive treatment).
Attacks against classification algorithms make a natural starting point since they are the most prominent example in the literature to date.
While our current experiments are purely digital (i.e. we do not physically realize AE), and therefore subject to certain limitations, they do provide useful insights about how AE for remote sensing algorithms may be generated under realistic physical constraints.
In \cref{sec:rs-considerations} we describe important considerations related to AE that arise in the remote sensing domain and how these considerations guide our problem setup.
%
%
%
%
\Cref{sec:adversarial-examples} reviews adversarial examples and outlines our optimization problem for constructing AE.
Finally, \cref{sec:experiments} provides numerical experiments in the context of a modern satellite imagery classification problem.
We conclude with suggestions for future work that would enrich the analysis of AE in this space.

\section{Approach} \label{sec:approach}
\subsection{Remote Sensing Considerations} \label{sec:rs-considerations}

In this section we describe salient aspects of the remote sensing problem which must be accounted for when considering a \emph{physical attack}, i.e. the creation of AE by manipulation of the physical world. The attack could be performed by altering the radiance or reflectance properties of the scene, e.g. by adding or removing materials or changing the surface texture. The exact mechanisms for altering the scene reflectance are outside the scope of this paper, though their consideration guides how the constrained AE problem is set up for this application.

\subsubsection{Viewpoint Geometry} \label{sec:viewpoint-geometry}
%
For near-nadir observations from a satellite in low-Earth orbit, variations in range to various objects in the scene are likely to be modest. Nevertheless, AE robustness to scale is an important issue and we will capture this source of variability in our experiments.
While the full three-dimensional geometry of the problem is quite relevant, we will assume for the purposes of this initial study that all objects lie entirely in the ground plane (i.e., the elevation above sea level for each pixel is zero) and that the off-nadir angle is modest. 
Therefore, any occlusions or shadows imposed by varying elevations of scene elements are ignored. 
%
With sufficiently high-resolution metadata providing the satellite ephemeris at the time the imagery was collected, future studies might entertain using an advanced projection/transformation of the images to the ground plane (where each pixel is mapped to a latitude/longitude/elevation).

%

We would observe, however, that the mild off-nadir angle assumption may not be totally egregious since, in many applications, large off-nadir angles may inherently confound image analysis due to the degradation in pixel resolution. 
Therefore, an adversary may be fundamentally  unmotivated to generate attacks for relatively extreme geometries.



\subsubsection{Atmospheric Effects}
Unlike machine learning algorithms applied to imagery taken from the surface of the Earth, the performance of algorithms in remote sensing applications is highly susceptible to environmental and atmospheric effects (e.g. illumination, clouds, haze) and the properties of AE are expected to be affected by these phenomena as well.

\subsubsection{Temporal Variability} \label{sec:temporal-variability}

\begin{figure}
\begin{minipage}[b]{.5\textwidth}
\centering
\includegraphics[width=2.4in]{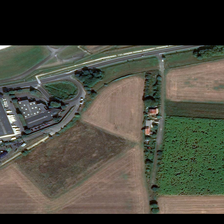}
\end{minipage}
\hfill
%
%
\begin{minipage}[b]{.5\textwidth}
\centering
\includegraphics[width=2.4in]{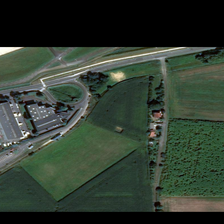}
\end{minipage}
\hfill
\caption{Scene with class label ``crop field'' exhibits substantial variability over time, due in large part to changes in ground vegetation.}
\label{fig:vegetation}
\end{figure}

Remote sensing data also has a temporal aspect that is distinct from video. 
An orbiting satellite's ground track will pass through the same spot on the Earth's surface according to a regular schedule (usually several days), enabling it to collect imagery over the same scene. 
For example, 
satellite systems such as Sentinel-1 image the entire Earth over a period of days which increases the importance of temporally-aware algorithms~\cite{zhu2017deep}.
However, the elements of the scene and the environmental conditions surrounding it may change between revisits (e.g. seasonal changes in vegetation, human patterns of life, and weather). 
For example, \cref{fig:vegetation} shows one example of how ground conditions can vary dramatically over time.
Algorithms that exploit remote sensing data, including AE, must be robust to changing conditions such as these.

In our experiments, we use a data set where the sensing system makes multiple but relatively infrequent passes over a given scene and we measure the effectiveness of an attack across the entire sequence.

\subsubsection{Physical Scale}
The scale of a satellite image is orders of magnitude larger than an image taken on the Earth's surface, and the number of pixels is typically much greater. 
For example, images taken from the IKONOS satellite can span up to 11.3 km across~\cite{digitalglobe2013ikonos} at sub-meter resolution. 
The large scale of these scenes imposes a significant challenge upon designing AE with so-called ``small'' perturbations, especially if they are to be designed in the physical realm. Approaches that subtly perturb entire images are implausible; instead, a practical AE will likely be restricted to perturbing a few regions of modest physical dimension.
Constraints on the physical support of AE have been considered already in the context of natural images and local sensing problems (e.g. \cite{brown2017adversarial,evtimov2017robust}); however, the potentially vast scale coupled with material constraints adds an subtle but important twist to this consideration.

\subsubsection{Material/Signature Properties}
Material and sensor properties also influence how an adversary may be able to manipulate a scene.
For example, in multi/hyperspectral imaging, one may not be able to arbitrarily modify the spectral signature of a given pixel.
Instead, material mixture models may determine the admissible set of perturbations that can be realized.
Alternatively, in multi-modal settings (e.g. LIDAR+EO), there are practical constraints upon how a subset of the scene can be modified jointly in each modality.
These challenges present opportunities to consider more sophisticated formulations for adversarial attacks.
For this foray into AE attacks on remote sensing, we will limit our experiments to the visual spectrum; however, note that that the data set we employ also includes multispectral signatures which could be used in future work.

\subsection{Adversarial Examples} \label{sec:adversarial-examples}
In this section we briefly review techniques for generating adversarial examples.
Space precludes an exhaustive review; additional details are available in~\cite{fawzi2017robustness} and the associated references.
In the remainder we will consider attacks against classification problems for image-like data, however it should be noted that the notion of AE generalizes to other settings as well.

%
Let $f : \R^d \rightarrow \{0, \ldots, k-1 \}$ denote a classifier which maps $d$-dimensional images to a discrete label set of cardinality $k$.
For a given input $x \in [0,1]^d$ (an image with pixels scaled to be between 0 and 1) and a target label $\ell \in \{0, \ldots, k-1\}$ let $r \in \R^d$ denote a perturbation designed to cause the classifier to predict label $\ell$; i.e. $f(x+r) = \ell$.  
We then refer to $r$ as an adversarial perturbation and the resulting signal $x+r$ as an adversarial example.
An underlying assumption is that $f(x) \neq \ell$ since it makes little sense to speak of an attack whose intent is to leave a classifier's original prediction unchanged.
The goal to induce a particular prediction $f(x+r) = \ell$ is termed a \emph{targeted attack}.
An alternative is a \emph{non-targeted} attack where the adversary's goal is merely to change the original prediction, i.e. $f(x+r) \neq f(x)$.  
If all details of the classifier $f$ are known by the adversary, this is often referred to as a \emph{white box}  attack.
When details of the classifier are not directly known and must be guessed or estimated via query, this is referred to as a \emph{black box} attack.
Generally speaking, white box attacks tend to be the most difficult to defend against~\cite{athalye2018robustness}.

The seminal paper on adversarial examples proposed the following  optimization problem~\cite{szegedy2013intriguing}:
\begin{align*}
    r^* = \min_r ||r||_2 \quad \text{subject to} \\
    f(x+r) = \ell \\
    x + r \in[0,1]^d.
\end{align*}
(The last constraint serves to ensure that the adversarial example $x+r$ has pixel values in the same range as $x$).
The authors also considered penalty function methods based on the classifier's loss function $J(x,y,\theta)$ 
%
%
%
\begin{align*}
    r^* = \min_r & \; c \|r\|_2  + J(x+r, \ell, \theta) \quad \text{subject to} \\
    & x + r \in [0,1]^d;
\end{align*}
here, $c$ is a scalar weighting coefficient and $\theta$ represents the network parameters/weights.
When training a neural network, $J$ is often taken to be a cross-entropy loss and $\theta$ is the design parameter; for adversarial attacks, the network weights $\theta$ are fixed and it is the perturbation $r$ which is optimized.

Subsequently a number of alternative approaches have been explored in the literature.  The fast gradient sign (FGS) method~\cite{goodfellow2014explaining} is a computationally thrifty heuristic for untargeted attacks subject to an $\ell_\infty$ constraint on the perturbation.
It uses the gradient of the classifier loss function $J$ to take a single step of magnitude $\epsilon$ in a direction that (locally) leads to the greatest increase in loss
\[
   r = \epsilon \, \text{sign} (\nabla_x J(x,y(x), \theta));
\]
here, $y$ represents a function that maps inputs to their true labels.
Iterative variants of FGS were presented in~\cite{kurakin2016adversarial}.  
Highly effective attacks for a variety of $p$-norm constraints upon the perturbation magnitude were presented in~\cite{carlini2017towards}.

The aforementioned methods all implicitly assume the adversary can manipulate all pixels in the scene.
Another interesting source of constraints arises when an attack can only modify a subset of the image support.
In~\cite{papernot2016limitations} the authors use a greedy iterative method to generate sparse perturbations that modify only a subset of pixels in the input space.
In~\cite{brown2017adversarial} the authors consider adding relatively modest-sized ``patches'' into natural image scenes in order to defeat whole-image classification algorithms.  
Analogously, \cite{evtimov2017robust,chen2018robust} seek to defeat street sign classification and detection by designing physical perturbations whose supports are explicitly constrained to coincide with that of the object in question (in these cases, restricted to the surface of a stop sign).
In these physical settings, the aforementioned authors seek to develop attacks that are effective over a range of possible locations, rotations, and scalings.
To achieve robustness to these possible variabilities, the authors leverage the idea of optimizing over a collection of potential transformations, an idea introduced by~\cite{athalye2017synthesizing} and termed \emph{expectation over transformation} (EOT). 
In this setting, the idea is to constrain the effective distance $\delta$ between adversarial and original inputs over a distribution of transformation functions $T$:
\begin{equation} \label{eqn:eot-delta}
\delta = \mathbb{E}_{t \sim T} \left[ d(t(x) - t(x+r)) \right],
\end{equation}
where $d$ is some suitable distance metric (e.g. a $p$-norm) and $t$ denotes the stochastic transformation.
The distribution $T$ can capture a variety of transformations, such as scaling, translation or additive noise.
Once a suitable $T$ has been determined, adversarial perturbations are generated by solving an optimization problem; the authors of~\cite{athalye2017synthesizing} proposed a problem of the form
\begin{equation} \label{eqn:eot}
 \underset{r}{\arg \min} \; \mathbb{E}_{t \sim T} [- \log \mathbf{P}(y| t(x+r))] + \lambda \, \delta,
\end{equation}
where $\mathbf{P}(y|x)$ is the classifier's probability estimate for label $y$ given image $x$ and $\lambda$ is a scalar weighting coefficient.

The EOT framework is quite general and therefore applicable to many settings. 
Of course, for attacks that are ultimately to be realized in the physical world, the question of designing $T$ so that it is sufficiently representative of the physical setting is non-trivial.
In most cases, precisely modeling all possible physical phenomena is unrealistic.
Thus, each new domain and sensing modality introduces new challenges and opportunities for analysis as one adapts $T$, the constraints, and the optimization procedure for the specific problem at hand.
%

\subsection{Digitally Emulating Physical Attacks}
Following in the spirit of \cite{brown2017adversarial}, we seek to develop opaque ``patches'' that, when applied to satellite imagery, defeat whole image classification algorithms.
Our goal is to design digital experiments that are, to the extent possible, representative of the physical setting where attacks would ultimately be realized.
The source signals we wish to attack occupy some physical space $\mathcal{P}$ whereas the sensed/observed inputs to a classifier have undergone a transform related to the sensing process.
We will refer to the sensed input space as $\mathcal{O}$ and model the sensing process by a function $\tobs : \mathcal{P} \rightarrow \mathcal{O}$ (here we are just giving explicit names to the spaces assumed in the EOT framework).

We do not assume access to a complete simulation of $\mathcal{P}$ and instead rely on sensed data to design our perturbations.  
For example, if $x \in \mathcal{P}$ is a source signal we wish to attack, we must do so by working with $\tobs(x) \in \mathcal{O}$.
Our process is to parameterize a physical perturbation $r \in \mathcal{P}$  and to then solve an optimization problem to design $r$ by computing $\tobs(r)$ and overlaying this ``sensed'' representation of $r$ into $\tobs(x)$.
Thus, the typical gradient descent procedure designs $r$ by backpropagation through $\tobs$.
In this approach, we assume that each pixel in $\mathcal{O}$ is associated with exclusively with either the patch-like attack or the original scene $\tobs(x)$ (i.e. no pixel receives contributions from both sources).  In doing so, we are essentially ignoring possible edge effects at the interface between the attack and the original signal.

Of course, $\tobs$ is subject to uncertainty and, as mentioned in \cref{sec:adversarial-examples}, the typical approach is to average over a distribution of transformation functions $T$. 
A complementary notion, which we explore here, is to use metadata associated with multiple observations $\tobs_1(x), \tobs_2(x), \ldots, \tobs_m(x)$ and approximate $T$ by means of these explicit samples from the true underlying distribution.

For example, ground sample distances provide explicit guidance in terms of how a perturbation $r$ (designed in physical coordinates) must be scaled so that $\tobs(r)$ is dimensionally consistent with $\tobs(x)$.
The quality of this approximation will of course depend upon complexity of the true sensing process and how much of that complexity the metadata permits one to incorporate into $\tobs$.

Another challenge with designing physical attacks in this setting is that, due to variations as described in section \ref{sec:temporal-variability}, the signal $x$ being attacked itself changes over time (and hence, from one observation to the next).
Thus, our data set reflect variations both due to the sensing process and the evolution of the targeted signal $\{x_1, \ldots, x_m\}$.
We are further assuming that the attacker is not adapting the perturbation $r$ over time but rather must design a single attack that is effective across all the variabilities manifested in $\{\tobs_1(x_1), \ldots, \tobs_m(x_m)\}$.
This is subtly different from most prior work in physical attacks where the dominant source of variability is associated exclusively with the sensing process.
We propose the following optimization for implementing targeted attacks against a sequence of observations
\begin{align} \label{eqn:our-attack}
r^* = & \underset{r}{\arg \min} \; \sum_{i=1}^m J(t_i(x_i) \oplus t_i(r), \ell, \theta)  \nonumber \\
& + \lambda \sum_{i=1}^m d(t_i(x_i) \oplus t_i(r)) \\
& \text{subject to} \quad t_i(x_i) \oplus t_i(r) \in [0,1]^d, \quad i =1, \ldots, m. \nonumber
\end{align}
The $\oplus$ in \cref{eqn:our-attack} is used to denote that we overlay the attack onto the sensed image (as opposed to component-wise addition, as is often done for subtle adversarial attacks).

Note that the observations $\{\tobs_i(x_i) \}_i$ used to design the perturbation need not consist of all samples present in the data set.
For example, one may be interested in exploring how well attacks designed using a subset of samples generalize.
%

The penalty term $d$ in \cref{eqn:our-attack} is another important design consideration.
In settings where visual subtlety is desired, one could follow the conventional approach of realizing $d$ using a suitable $p$-norm.  
In this case, this would be asking for a perturbation that is generally subtle across a range of observational conditions and variations in $\{x_i\}$.
The feasibility of visual subtlety is therefore somewhat signal dependent.

Ideally, attacks would also be subtle with respect to the 3D geometry of the scene.
For example, if the spatial extent of the patch extends beyond the roof of a building or spans regions that are partially obscured/in shadow, it would be ideal if the perturbation $r$ respected these discontinuities.  
In the absence of a full 3D model of $\mathcal{P}$ 
another approach would be to segment the scenes $\tobs(x)$ and incorporate this structure into the attack model.
In our current work we loosely approximate this by adding a second  term to $d$ which encourages the attack to respect strong edge structure within an image (as determined by a Canny edge detector).


Obviously there is opportunity for designing both $\tobs$ and $d$ to make attacks increasingly realistic.
Of course, complexity should not be added arbitrarily to these terms since one must also consider impact to the feasibility of solving the resulting optimization problem (e.g. gradient based methods will require reasonably tractable and benign functions $\nabla_r \tobs, \nabla_r d$).


\section{Numerical Experiments} \label{sec:experiments}
\subsection{Data Set}

We base our numerical experiments on the Functional Map of the World (fMoW) data set, a large collection of temporal sequences of satellite imagery designed for evaluating whole image classification problems related to the functional purpose of land use and buildings~\cite{christie2017functional}.
The overall fMoW data set contains over a million images from 200 countries; our experiments utilize a subset of the validation split (the overall validation split contains approximately 53000 images from 12000 unique scenes). 
Each image contains at least one bounding box labeled with one of 63 possible classes.
Furthermore, the data set includes metadata features pertaining to location, time, sun geometry, cloud cover, and physical dimensions of the imaged swath.
The fMoW data set has multiple modalities: 4-band or 8-band multispectral imagery as well as high and low resolution RGB imagery.  
For  our experiments, we use the high resolution RGB images but observe that the multispectral domain provides a compelling setting for future study.

\begin{figure}
\centering
\includegraphics[width=4in]{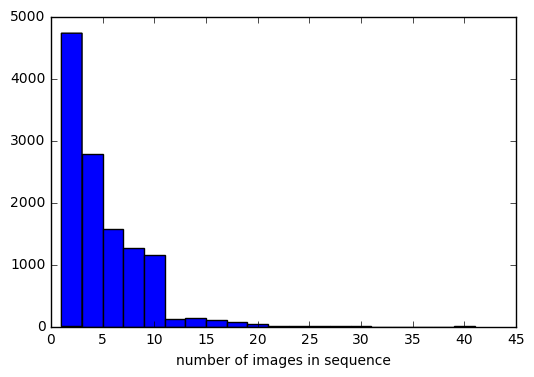}
\caption{Number of (RGB) images per sequence in the fMoW validation split (median=3, maximum=41).}
\label{fig:images-per-sequence}
\end{figure}

We further downsampled the validation subset of fMoW so the resulting sequences satisfy a number of desiderata.
First, since our interest is in attacking sequences of nontrivial length, we only consider those with at least 8 views of the same scene.
This eliminates a large number of sequences as many in the validation set are quite short (see \cref{fig:images-per-sequence}).
We also only consider images that are correctly classified by the targeted classifier prior to applying any perturbation.
Finally, we also limit our study to relatively benign sensing conditions (recall the discussion from \cref{sec:viewpoint-geometry}).
We only use images with: mild off-nadir angle (less than 30 degrees), at most modest cloud cover (less than 20 percent of image chip obscured), and sun elevation angles of at least 60 degrees to eliminate more extreme variations in illumination.
The resulting experiment includes 66 sequences each having at least 8 admissible frames.

The authors of fMoW also developed and analyzed a number of classification models, some of which use exclusively image data while others make use of metadata and/or exploit the sequential nature of the images by using recurrent networks (e.g. LSTMs). 
Our study considers the one designated ``CNN-I'', which is a fine-tuned variant of DenseNet with no recurrent structure designed for RGB image data.
Our choice of CNN-I is justified in this case since the performance of this network is quite close to that of the recurrent alternatives (see table 1 in \cite{christie2017functional} for more details).
However, extending the analysis to other networks (e.g. those using metadata or that more explicitly incorporate temporal properties) is another interesting direction for future work.

The fMoW authors also make available their preprocessing algorithm which extracts and resizes bounding boxes into a tensors of dimension 229x229x3 suitable for ingestion by CNN-I/DenseNet.
This rescaling has implications for the associated metatdata. 
In our study we adopt the fMoW preprocessing strategy so that we can use CNN-I without further modification; however, as part of the preprocessing we  also update the metadata so that salient values (e.g. ground sample distance) are still representative following the rescaling.

There remain a few challenges associated with this data set, however.
One substantial issue is that the images for a given scene are not precisely registered.  
Therefore, while we can use metadata to properly control the scale of a patch, there is no guarantee that the attack will be translated to the precise same location from one observation to the next.  
Thus, our attack will be subjected to some positional uncertainty.

\subsection{Algorithm Implementation} \label{sec:algo-implementation}

For our experiments, we selected 4 target classes from the fMoW taxonomy: ``crop field'', ``hospital'', ``office building'', and ``park''.  
We do not try every possible targeted attack only in order to control the computational expense of our experiments; these four classes were selected so as to have some representation of both urban and rural scenery.
This selection was made prior to algorithm evaluation and is therefore not cherry-picked for performance.

Attacks are modeled as a single, opaque, flat, piecewise-constant square surface consisting of $n \times n$ elements.
Each element is associated with a single RGB value that is tuned during optimization.
For each image, the attack is placed in the center of the scene.
The number of elements $n$ and the size, in meters, of each element are experimental parameters that are fixed prior to learning the attack.

We implemented our experiments in Python using the TensorFlow \cite{tensorflow2015-whitepaper} machine learning library.
We use the typical cross-entropy loss for $J$. 
The function $\tobs$ uses the metadata (in particular, the ground sample distance) to scale the attack patch appropriately for each image.
The penalty function $d$ has two terms: one which encourages a small $\ell_2$ distance between the attack and the underlying images while the second term encourages a small $\ell_2$ distance between the attack and strong edge-like structure in the scene.
These two terms have an associated weighting coefficients of $\lambda_1 = 1\mathrm{e}{-3}, \lambda_2=1\mathrm{e}{-1}$.
We used the Canny edge detector provided as part of skimage to compute edge structure (with a Gaussian width parameter of 2.0) for all images.
Calculations were performed on a GeForce 1080 Ti trained using gradient descent for 1000 epochs at two different learning rates (100 and 20) which took approximately 10 minutes per attack (although we observe that our code was not optimized to make use of the most efficient possible CPU to GPU data transfers).

\subsection{Results}
Overall attack success rates for six different experiments (covering three different physical attack configurations) are shown in \cref{tbl:success-rates}.
The first three rows provide a baseline result for when an attack is based solely on the first image in a sequence.
Numbers reported are misclassification rates post-attack (note that all images used in this study were correctly classified by CNN-I pre-attack).
The second three rows show how the overall attack rate improves if the attacker is privy to the first four images in each sequence.

\begin{figure}[t]
\centering
\includegraphics[width=3.25in]{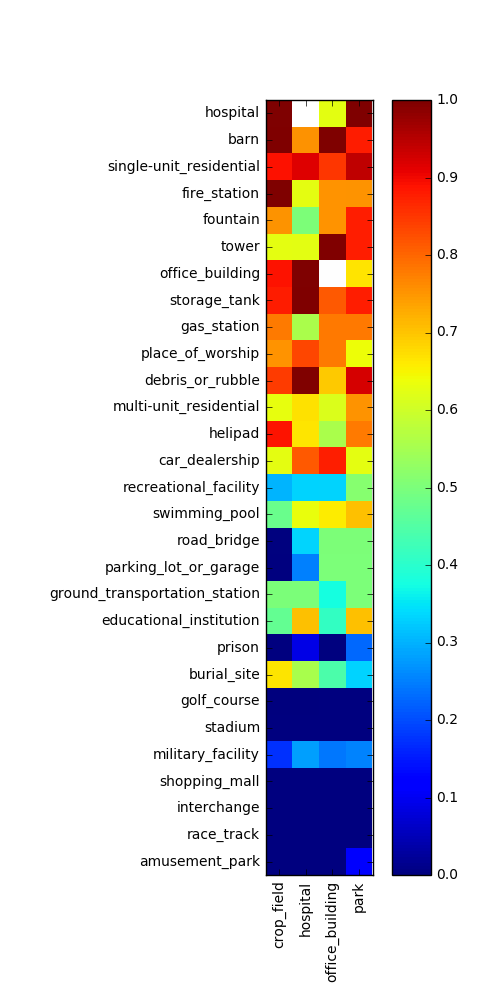}  
\caption{Targeted attack success rates by sequence label (vertical axis) and target label (horizontal axis).  Results correspond to experiment 6 in \cref{tbl:success-rates}.
Original sequence labels are ordered by decreasing size, in pixels, of attacks leveled against images associated with that class.  Since attacks are all of the same physical dimension, this ordering also reflects the ground sample distances associated with images from these classes.}
\label{fig:confusion6} 
\end{figure}

\begin{figure}
\centering
\begin{minipage}[b]{.4\textwidth}
\centering
\includegraphics[width=3in]{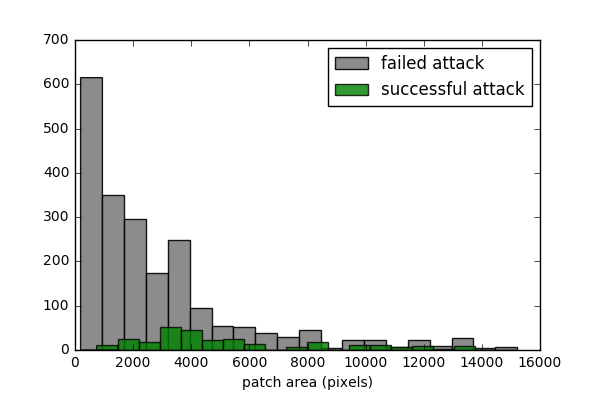}
\subcaption{experiment 1, targeted attacks}
\end{minipage}
\qquad
\begin{minipage}[b]{.4\textwidth}
\centering
\includegraphics[width=3in]{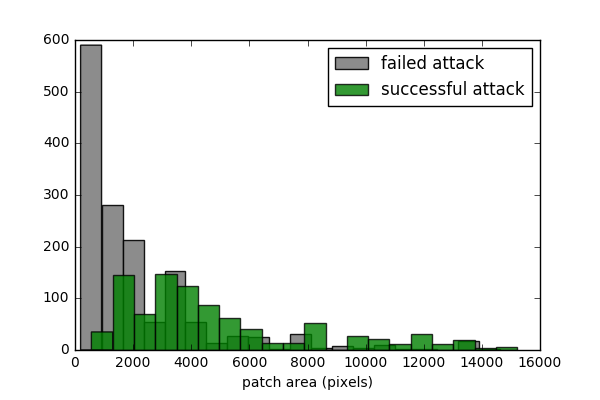}
\subcaption{experiment 1, non-targeted attacks}
\end{minipage}
%
\\
\begin{minipage}[b]{.4\textwidth}
\centering
\includegraphics[width=3in]{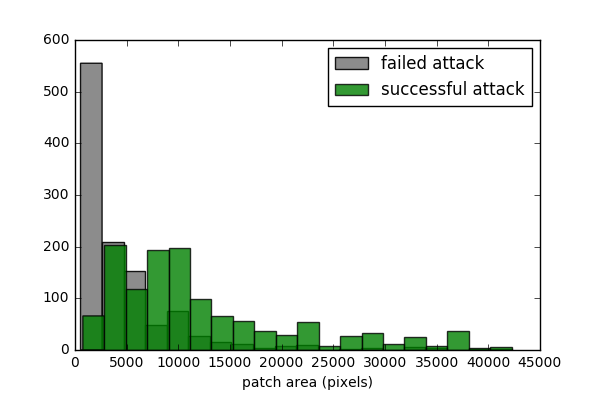}
\subcaption{experiment 6, targeted attacks}
\end{minipage}
\qquad
\begin{minipage}[b]{.4\textwidth}
\centering
\includegraphics[width=3in]{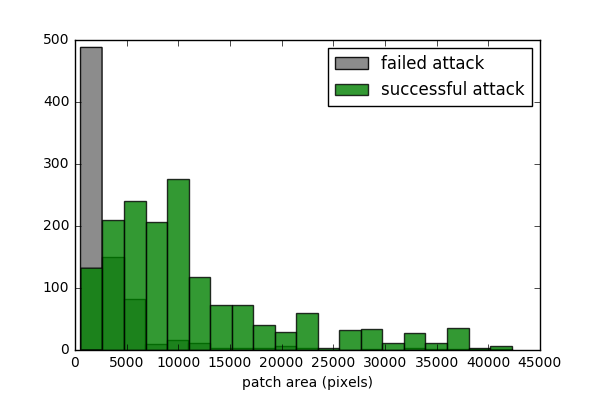}
\subcaption{experiment 6, non-targeted attacks}
\end{minipage}
\hfill
\caption{Distributions of successful and unsuccessful attacks as a function of number of pixels the attack manipulated.
Attack success rates increase as the attack is able to manipulate more pixels in the scene (here, corresponds to decreasing ground sample distances).}
\label{fig:success-vs-size}
\end{figure}

\begin{figure}
\centering
\includegraphics[width=2.5in]{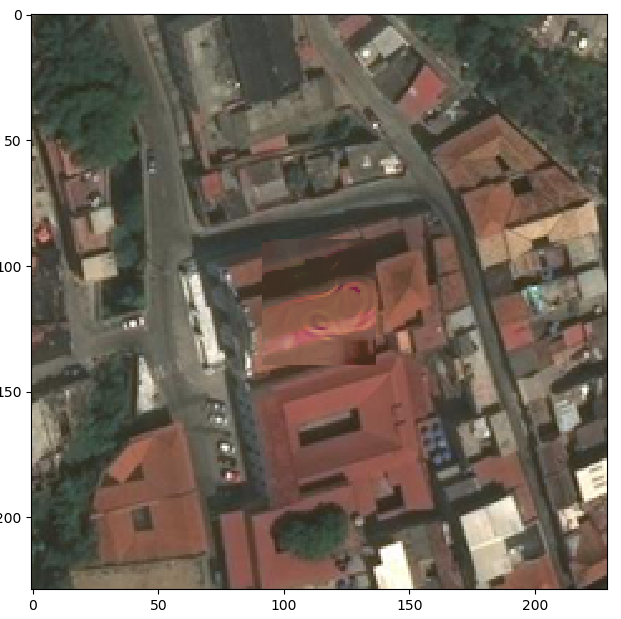}
\caption{Targeted attack causing the classifier to label a "place of worship" as a "hospital".  
The patch-like attack is located in the center of the scene.
  While the attack does not precisely respect the underlying 3D geometry, the edge/shadow constraints
  encourage partial agreement with the boundaries of the structure and the overall visual effect is reasonably subtle. }
\label{fig:shadow-example}
\end{figure}

\begin{figure}
\centering
\includegraphics[width=5in]{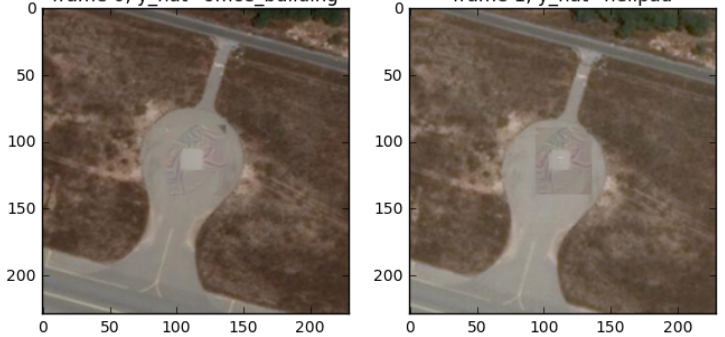}
\caption{Attack on first element of a "helipad" sequence (left) causes the classifier to label the scene "office building" (an unsuccessful targeted attack, but one that does induce a misclassification). The same attack fails to confuse the classifier in the second image of the scene (right), which  is correctly classified as "helipad".  
While the two images are quite similar, the subtle differences are sufficient to impact the effectiveness of this particular attack.}
\label{fig:helipad-example}
\end{figure}

\Cref{fig:confusion6} shows, in the case of experiment 6, targeted attack success rates broken down by class label.  
From this figure it is clear that some classes are more readily attacked in our experiments.
Note that many of the classes for which attacks are not successful (e.g. stadium, shopping mall, amusement park) might be expected to correspond with larger scenes  (images with greater ground sample distances).
Given that all images are rescaled to a fixed size by the fMoW preprocessing and that our attacks are designed in physical coordinates, larger scenes would result in fewer available pixels for the attack to manipulate.
The hypothesis that limiting the number of pixels available for an attack may reduce the success rate is further supported by \cref{fig:success-vs-size} where
we overlay distributions for successful and unsuccessful attacks as a function of number of pixels manipulated.
\Cref{fig:shadow-example} shows a visual example of how matching strong edges and shadows is encouraged by our choice of $d$; also clear, is that this does not provide perfect agreement with the 3D geometry of the scene.
As mentioned previously (\cref{sec:viewpoint-geometry}), if additional metadata were available the constraints on the spatial support of the attack could be made more realistic.
Similarly, \cref{fig:helipad-example} shows two frames from a scene which highlights how lack of perfectly registered images leads to some inconsistency in patch translation.
Improving upon these details is a ripe direction for future work.

\begin{table*}
\centering
\begin{tabular}{c|ccc|cc}  
& \multicolumn{3}{c|}{Attack Parameters} & \multicolumn{2}{c}{Metrics} \\
\hline
  &  &           &  num. frames  &  attack success  & total error      \\
experiment  &  $n$ & m/element   &  attacked     &  rate (\%)      & rate (\%)  \\ 
\hline \hline
1 & 60 & 0.5 & 1 & 11.9 & 38.1 \\
2 & 80 & 0.5 & 1 & 15.5 & 43.2 \\
3 & 100 & 0.5 & 1 & 19.2 & 50.8 \\
\hline
4 & 60 & 0.5 & 4 & 31.3 & 51.7 \\
5 & 80 & 0.5 & 4 & 44.5 & 61.5 \\
6 & 100 & 0.5 & 4 & 53.1 & 67.3 \\
\end{tabular}
\caption{Success rates for targeted white box attacks against the fMoW baseline classifier CNN-I.  
Attack parameters include the number of elements $n$ in each dimension as well as the size, in meters, of each element (see \cref{sec:algo-implementation}).
Also reported are frequency of classifier errors, independent of whether the target classification was induced (last column).  All unperturbed (i.e. original, non-AE) images were correctly classified.}
\label{tbl:success-rates}
\end{table*}

\section{Conclusion and Future Work}   \label{sec:future-work}
In this paper we have presented the first (as far as the authors are
aware) experiments in adversarial examples for satellite imagery classification.
We describe an approach for simulating physical attacks in the
digital space which is unique in our use of metadata to align the attack with remotely sensed data.
We also provide preliminary experiments indicating that, under certain assumptions, it may be possible to induce errors in state-of-the-art whole image classifiers for satellite imagery.

Of course, further refinements of these experiments (e.g. more realistic simulations of the sensing process, more extensive use of metadata, experimentation on larger sets of data, etc.) are all near-term next steps which would further enrich the results presented here.
Additionally, there is abundant room to explore the broader space of attack designs, which could include implementing multiple patch-like attacks (and the corresponding patch location design problem), more directly accounting for 2D and 3D structure of scenes (e.g. by combining attacks with segmentation results), incorporating different notions of visual subtlety, and also exploring the impact of different levels of knowledge about the targeted system on the part of the adversary.
More ambitiously, there are interesting questions regarding extensions to other modalities (multi and hyperspectral data, SAR, LIDAR, etc.) and also to settings where multiple sensors are utilized simultaneously.
These directions also offer new opportunities for defining relevant constraints, such as exploiting spectral signature databases and  material mixture models to ensure physically realizable perturbations in multi/hyperspectral settings.
Attacks to other signal processing algorithms in the remote sensing domain, such as object detection and change detection, also offer interesting opportunities.
More generally, as the field of adversarial examples continues to mature, new findings and discoveries may play a role in this setting as well.
As mathematical techniques for developing classifiers with provable desirable properties evolve (e.g. \cite{bruna2013invariant}) their findings may help inform future experiments of robustness in the remote sensing setting.
Finally, there is an obvious need to validate digital experiments physically.
Our hope is that this work provides a compelling starting point for future studies of adversarial examples in the remote sensing domain.

\ifACM
\else
  \bibliographystyle{plain}
\fi
\bibliography{refs}

\end{document}